\documentclass[letterpaper, 10 pt, conference]{ieeeconf}  

\IEEEoverridecommandlockouts                              
\usepackage{graphicx}
\usepackage{subfigure}
\usepackage{lipsum}
\usepackage{cite}
\usepackage{amsmath}
\usepackage{fancyhdr}
\usepackage{color}
\usepackage{multirow}
\usepackage{multicol}
\usepackage{wrapfig}

\graphicspath{
    {./fig/}
}
    
\title{\LARGE \bf
Support Generation for Robot-Assisted 3D Printing with Curved Layers
}

\author{Tianyu Zhang$^{1}$, Yuming Huang$^{1}$, Piotr Kukulski$^{1}$, Neelotpal Dutta$^{1}$, Guoxin Fang$^{1}$ and Charlie C.L. Wang$^{1\dagger}$
\thanks{$^{1}$All authors are with the Department of Mechanical, Aerospace and Civil Engineering, University of Manchester, Manchester.}%
\thanks{$^\dagger$Corresponding Author: {\tt\footnotesize changling.wang@manchester.ac.uk}}
\thanks{This work was partially supported by the chair professorship fund of the University of Manchester.}
}

\usepackage{color}

\begin{document}

\maketitle
\begin{abstract}
Robot-assisted 3D printing has drawn a lot of attention by its capability to fabricate curved layers that are optimized according to different objectives. However, the support generation algorithm based on a fixed printing direction for planar layers cannot be directly applied for curved layers as the orientation of material accumulation is dynamically varied. In this paper, we propose a skeleton-based support generation method for robot-assisted 3D printing with curved layers. The support is represented as an implicit solid so that the problems of numerical robustness can be effectively avoided. The effectiveness of our algorithm is verified on a dual-material printing platform that consists of a robotic arm and a newly designed dual-material extruder. Experiments have been successfully conducted on our system to fabricate a variety of freeform models. 
\end{abstract}


\section{Introduction}\label{secIntro}
Additive manufacturing (also called 3D printing) has radically changed the ways that products are made; meanwhile, robotic arms have been widely used in 3D printing (e.g.,~\cite{Bhatt_ADDMA20_survey, Urhal_19RCIM}). Compared to 
conventional 3D printers that use fixed printing direction, real 3D printing can be realized to fabricate curved layers of materials with the help of the additional \textit{degrees-of-freedom} (DOFs) provided by a robotic arm. 
Specifically, multi-axis 3D printing can reduce the need for supporting structures~\cite{Dai_SIG18}, enhance the mechanical strength of printed models~\cite{Fang_SIG20}, and improve the surface quality~\cite{Etienne_SIG19} by different curved layers optimized for different purposes. When trying to achieve the best performance in objectives rather than support-free, supporting structures (shortly called \textit{support} in the rest of this paper) are still needed for 3D printed curved layers. 
In this paper, we propose a skeleton-based support generation method for robot-assisted 3D printing with curved layers.

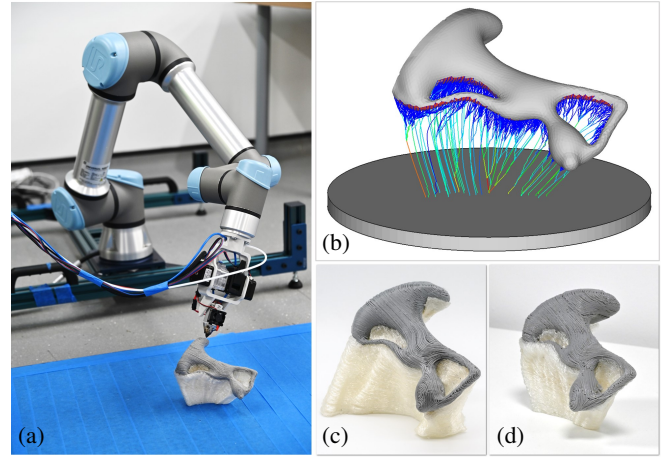
\begin{figure}[t]
\centering
\includegraphics[width=\linewidth]{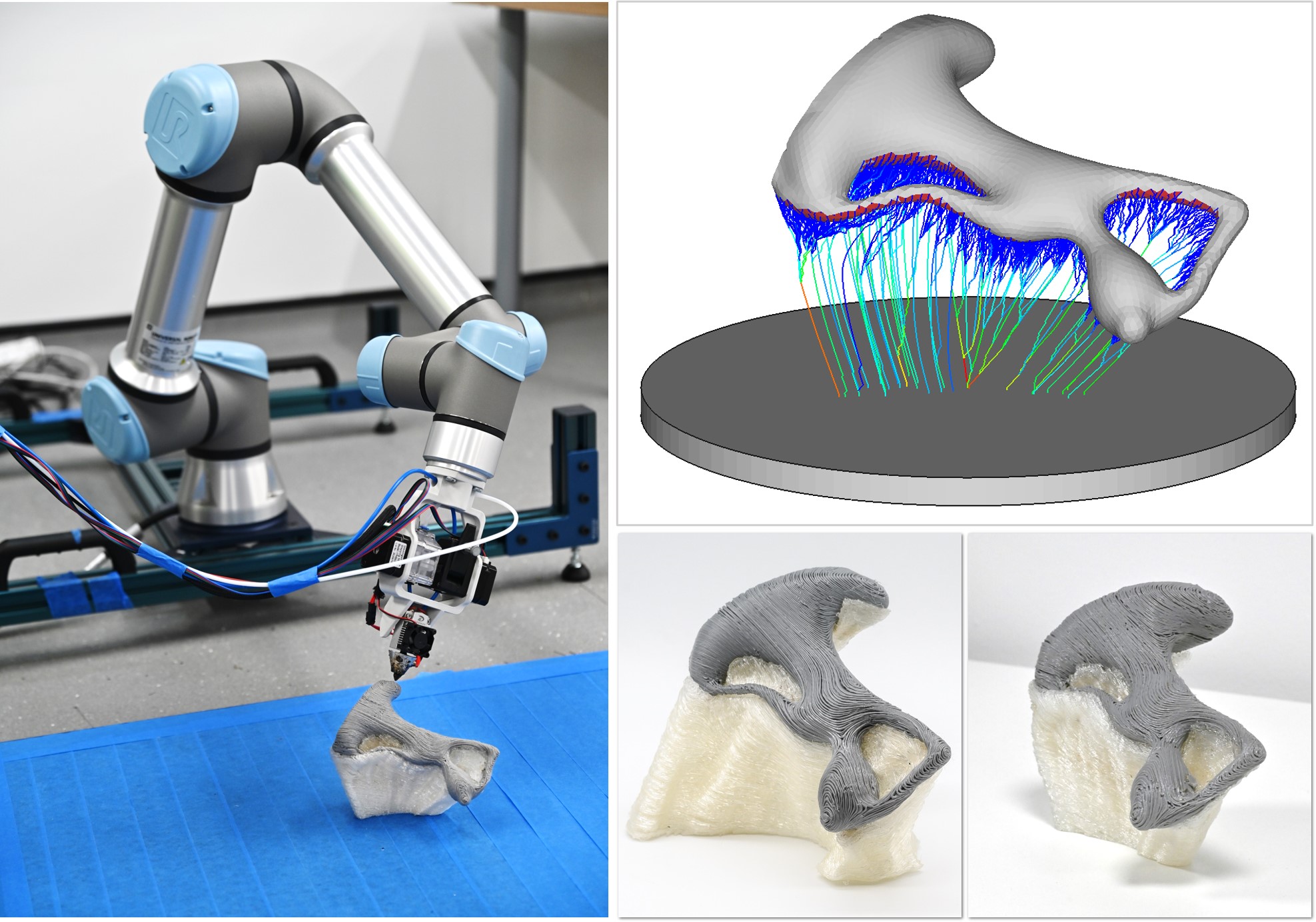}
\put(-243,5){\small \color{black}(a)}
\put(-128,78){\small \color{black}(b)}
\put(-128,5){\small \color{black}(c)}
\put(-62,5){\small \color{black}(d)}
\vspace{-5pt}
\caption{The robot-assisted 3D printing system and its results of fabrication: (a) a UR5e robot arm equipped with a dual-material extruder, (b) tree-skeleton of the support structure for the input model Yoga with its curved layers generated for reinforcing the mechanical strength~\cite{Fang_SIG20}, 
(c) the printing result using the support generated by the method presented in~\cite{Fang_SIG20}, and (d) the fabrication result by the support generation of this paper -- the volume of support has significantly reduced by 43.6\% and the total printing time is reduced from 13.8h to 9.1h.}
\label{figTeaser}
\vspace{-15pt}
\end{figure}

\subsection{Motivation}
As an essential part of conventional 3D printing systems with planar layers, support generation algorithms always detect the regions with overhang according to a fixed printing direction \cite{Huang_BOOK14}. The supports are added vertically below the overhang. However, the printing orientations for curved layers are dynamically varied along the toolpaths. As a result, the overhang detection and the support generation algorithm for plane-based printing cannot be applied. 

Few algorithms have been developed to generate support structures for curved layers in 3D printing. An algorithm was introduced in~\cite{Fang_SIG20} by first generating rays of sample points along their inverted surface normals, computing the $\alpha$-shape of the points sampled on the rays, and then trimming the curved mesh layers (obtained from the extrapolation of the field for generating curved layers in the enveloped solid) by the $\alpha$-shape. This approach mainly suffers from two issues: 1) the support does not have a compact volume (e.g., see Fig.\ref{figTeaser}(c)) and 2) the trimming step realized by Boolean operations on mesh surfaces is prone to problems of numerical robustness~\cite{Hoffmann_JCISE01,Wang_TVCG13}. The work proposed in this paper aims at solving these two problems while using the same framework for generating the curved layers for both the input solids and the supporting structures. Our algorithm is based on generating a tree-like skeleton and converting it into an implicit solid to trim the curved layers for support, which is more robust. 

We have tested the proposed algorithm on a robot-assisted 3D printing hardware as shown in Fig.\ref{figTeaser}(a). Contrary to the systems employed in \cite{Dai_SIG18,Wu_ICRA17}, this system incorporates a printer head mounted on the end effector of a UR5e robotic arm. A 2-in-1 design is developed for the extruder to support the printing of dual materials, which provides an extrusion solution that is more compact than the hardware setup used in \cite{Fang_SIG20}. Note that, as a general support generation algorithm for 3D printing with curved layers, the algorithm developed in this work can also be applied to the hardware with Cartesian motion and tilting table/head (e.g., those discussed in \cite{Zhang_21RAL}).



\subsection{Related work}
Planar layer-based 3D printing has been widely used in different fields such as printing metal for an aerospace turbine~\cite{Vafadar_AS21}, tooth alignment treatments in the medical field~\cite{Sangeeth_POLYMERS21} and printing of lattice structures for sportswear shoes~\cite{Vanderploeg_IJFDTE17}. Traditional planar printing is limited to three-axis movement with step motion along the Z-axis. This method of 2.5D printing is easy to implement and thus popular in the consumer market. However, the issues of low mechanical strength between layers, stair-case artifacts, and the requirement of support in large volumes are the generally discussed drawback of plane-based printing. For example, Wulle et al.~\cite{Wulle_CIRP17} analyzed such limitations of the current 3D printing method and proposed that \textit{multi-axis additive manufacturing} (MAAM) can enable new design and optimization possibilities than the conventional AM. Hence, more and more research works start to focus on multi-axis 3D printing. 

Robot-assisted printing platforms could provide extra DOFs and larger working space, and they also show very excellent mobility. Researchers in~\cite{Dai_SIG18, Li_CAD21,Wu_ICRA17} built multi-axis FDM printers for support-less or even support-free printing, which are composed of a 6DOFs robot arm (UR) and a fixed filament extruder. The robot holds the platform and the workpiece to achieve desired poses with respect to the nozzle of a fixed printer head. The advantage of this method is its high versatility. Because different tools can be pre-installed on the frame, there is no need to worry about filament or wire winding and avoid the trouble of constantly changing tools. But in order to effectively use this setup,  the transformation matrix from tool to platform center needs to be calibrated accurately, and it is challenging to get an accurate calibration in the whole working space of 3D printing that has large regions away from the platform of 3D printing. Contrary to this configuration, Soler et al.~\cite{Soler_ACADIA17} changed the orientations of the extruder during fabrication while fixing the object to be printed, which is suitable for printing large objects. Two robot arms were arranged to construct a whole system in~\cite{Bhatt_ADDMA20} where one robot is used to rotate the platform and another is used to move the extruder to build thin shell parts. 
A robot-assisted system that couples a 6-DOF robotic arm with an additional 2-axis tilting and rotatory table was developed in~\cite{Ding_RCIM17,Fang_SIG20}. 
Moreover, Fang et al.~\cite{Fang_SIG20} also provided an $\alpha$-hull based support generation method for overhang regions~\cite{Evans_BOOK12} for 3D printing with curved layers. More research on robot-assisted additive manufacturing can be found in other comprehensive literature reviews such as~\cite{Bhatt_ADDMA20_survey}.


In order to print the models with overhang regions, Vaissier et al.~\cite{Vaissier_CAD19} first used a lattice cell to fill the space of support as an initial guess, and then removed unnecessary lattice beams to obtain tree-like support. Dumas et al.~\cite{Dumas_SIG14} took advantage of the ability to bridge the gap of FDM and proposed a scaffold structure, which is composed of bridges and vertical pillars to support the overhang regions. This method can provide more stable support for printing compared with tree-like support. The self-supporting cone is used to decrease the usage of support material in~\cite{Schmidt_SIG14}. A new type of tree-like support, named \textit{Escaping Tree-Support} (ET-Sup) was proposed in~\cite{Kwok_RPJ21} to build all the supports onto the building platform to minimize the number of contact points. However, all the existing works only focus on the support generation for planar layer-based printing. Few works have been investigated for support generation for 3D printing with curved layers.

\subsection{Contribution}
Our major contribution can be summarized as follows:
\begin{itemize}
    \item A skeleton-based support generation algorithm for 3D printing with curved layers, which can output supporting structures with more compact volumes.
    
    \item The supports are represented as implicit solids so that the problems of numerical robustness in Boolean operations can be effectively avoided.

\end{itemize}
This is an important and essential extension of our previous work~\cite{Fang_SIG20}. The effectiveness of this new algorithm has been tested and verified on the robot-assisted hardware which is equipped with a newly designed 2-in-1 extruder.

\section{Framework and Overview}\label{secSystem}
\subsection{Field-based generation of curved layers}
Our approach was developed in the framework of a field-based slicing algorithm for generating curved layers for multi-axis 3D printing~\cite{Fang_SIG20}. The input of this framework is a model represented by a tetrahedral mesh $\mathcal{T}^m$. Depending on chosen objectives (e.g., the mechanical strength reinforcement), a vector field $\mathbf{V}(e)$ is computed on $\mathcal{T}^m$ and the optimized vector $\hat{\mathbf{v}}_e$ defined at each element $e$ indicates the local printing direction $\mathbf{d}_p$ inside $e$. Then, the governing field $G(\mathbf{x})$ with field values defined on nodes can be computed by solving the following minimization problem
\begin{equation}
G(\mathbf{x}) = \arg \min \sum_{e \in \mathcal{T}^m} \| \nabla G(\mathbf{x}_{e}) -  \hat{\mathbf{v}}_e \|^2, 
\end{equation}
where the gradient $\nabla G(\mathbf{x}_{e})$ is in the form of a linear combination of field values defined on the four nodes of an element $e$ -- details can be found in~\cite{Fang_SIG20}. Finally, a sequence of iso-surfaces $\{\mathcal{L}^m_i\}_{i=1,2,\cdots,n}$ is extracted from $G(\mathbf{x})$ to be used as the curved layers for multi-axis 3D printing.

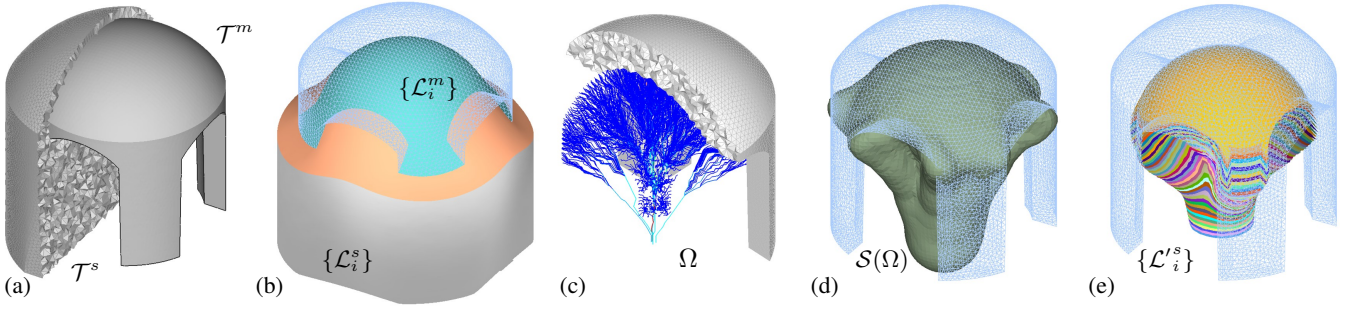
\begin{figure*}[t]
\centering 
\includegraphics[width=\linewidth]{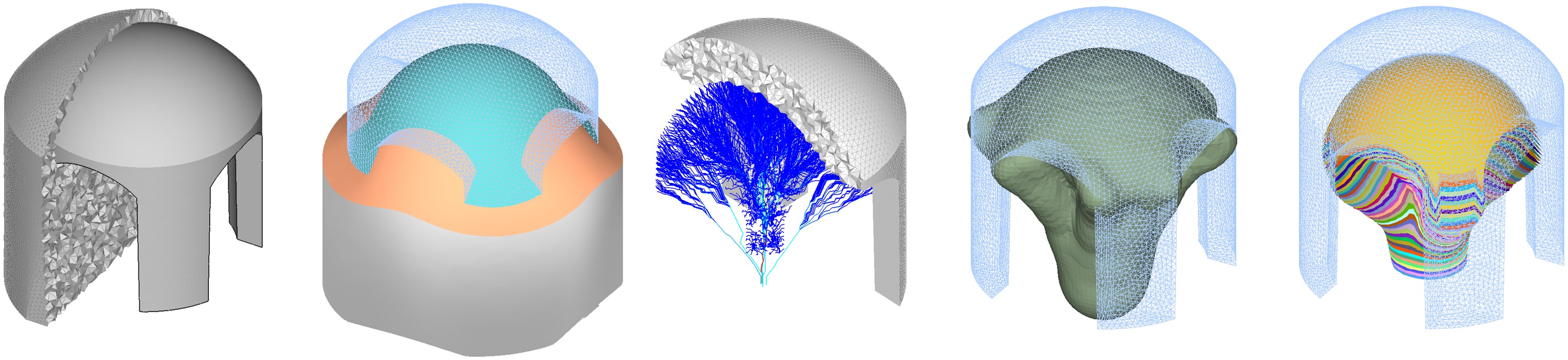}
\put(-505,5){\small \color{black}(a)}
\put(-410,5){\small \color{black}(b)}
\put(-295,5){\small \color{black}(c)}
\put(-200,5){\small \color{black}(d)}
\put(-95,5){\small \color{black}(e)}
\put(-425,100){\small \color{black}$\mathcal{T}^m$}
\put(-480,10){\small \color{black}$\mathcal{T}^s$}
\put(-356,80){\small \color{black}$\{\mathcal{L}^m_i$\} }
\put(-385,15){\small \color{black}$\{\mathcal{L}^s_i\}$}
\put(-250,15){\normalsize \color{black}$\Omega$}
\put(-183,15){\small \color{black}$\mathcal{S}(\Omega)$}
\put(-77,15){\small \color{black}$\{\mathcal{L'}^s_i\}$}
\caption{Diagram to show the pipeline of our support generation algorithm for 3D printing of curved layers. (a) Computational domain of support structure; (b) Compatible layer set; (c) Support tree skeleton; (d) Implicit surface of tree skeleton; (e) Slimmed support layers after extraction.}\label{fig:sys_pipeline}
\vspace{-10pt}
\end{figure*}

\subsection{Computational domain and layers for supports}
First, we determine the computation domain as the envelope in which supports need to be added. On the surface of a given model, the overhang regions are those satisfying ~\cite{Hu_CAD15}
\begin{equation}\label{overhang detection}
    \mathbf{n}_f \cdot \mathbf{d}_p + \sin(\alpha) \leq 0,
\end{equation}
where $\mathbf{n}_f$ is the normal of a boundary face of tetrahedral mesh $\mathcal{T}^m$. Instead of \cite{Hu_CAD15} that using a fixed local printing direction $\mathbf{d}_p$, 3D printing of curved layers has different $\mathbf{d}_p$ in different regions. A set of overhang faces $\mathcal{F}_o$ can be found from the boundary surface of $\mathcal{T}^m$ according to the varied $\mathbf{d}_p$s. Specifically, support-free is achieved when $\mathbf{d}_p$ falls into the cone shape determined by $\mathbf{n}_f$ and the self-supporting angle $\alpha$ which commonly depends on the printing material, temperature, and the nozzle size. $\alpha = 45^\circ$ is used in our experiments. 

\begin{figure}[t]
\centering 
\includegraphics[width=\linewidth]{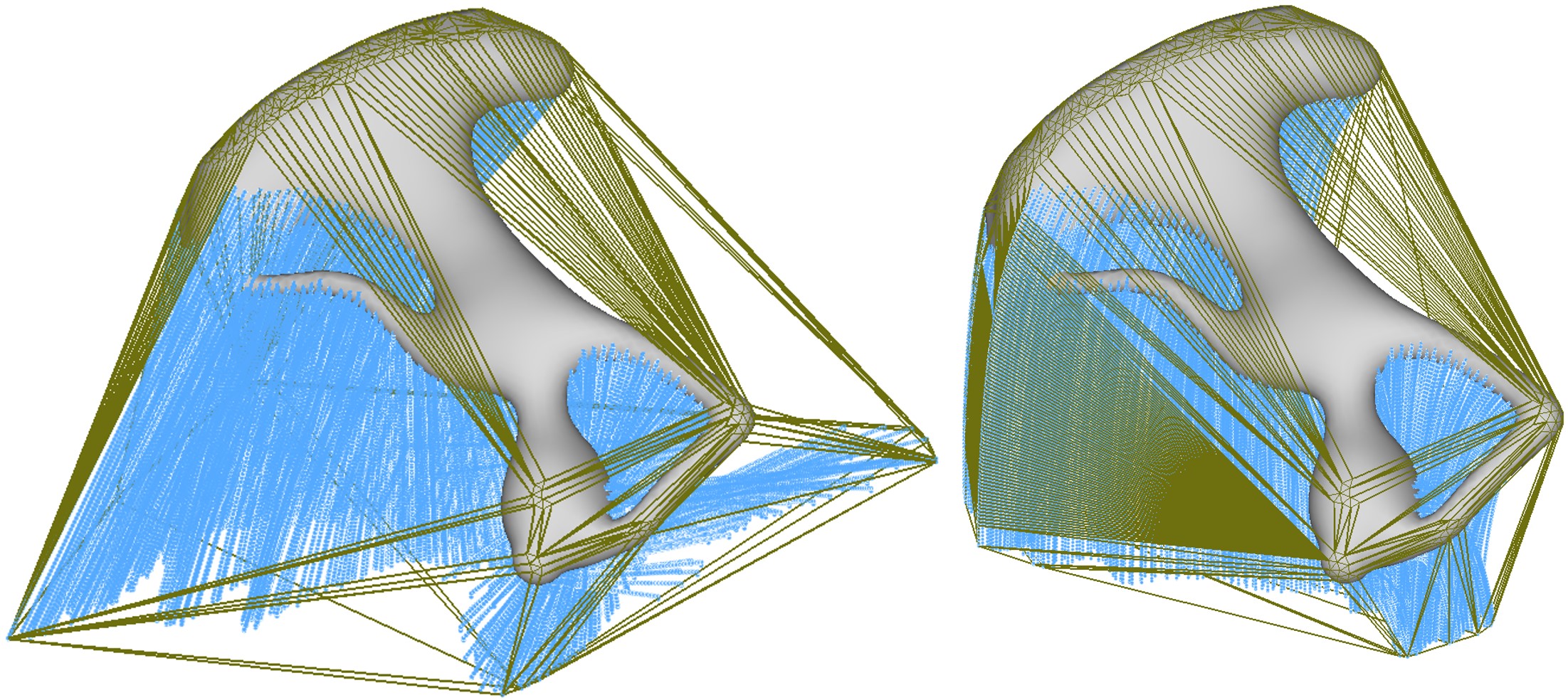}
\put(-245,0){\small \color{black}(a)}
\put(-100,0){\small \color{black}(b)}
\caption{The conservative hull for building the computational domain. (a) The strategy used in \cite{Fang_SIG20}; (b) The current method which avoids the intersection of the line segment with the bottom very far from the origin when $\mathbf{d}_p$ is almost parallel to the bottom.}
\label{fig:CompHull}
\vspace{-10pt}
\end{figure}

First, we apply a conservative strategy to determine the computational domain of support generation as $\mathcal{T}^s$. The curved support layers, that are compatible with the curved layers in $\mathcal{T}^m$, can then be computed in $\mathcal{T}^s$. For each overhang face $f \in \mathcal{F}_o$, we `project' every vertex $\mathbf{p}$ of $f$ as a particle towards the building platform $\mathcal{P}$ (i.e., a plane perpendicular to $z$-axis) along a trajectory determined by the following steps: 
\begin{itemize}
\item Starting from a direction $\mathbf{u} = -\mathbf{d}_p$, we progressively move $\mathbf{p}$ to a new position as $\mathbf{p}+ d \mathbf{u}$ with $d$ being the desired thickness of each layer. 

\item After each step of movement, we turn the moving direction $\mathbf{u}$ towards $(0,0,-1)$ with an angle equal to $\alpha/20.0$ until $\mathbf{u}=(0,0,-1)$. This $\alpha/20.0$ is decided by experiment to ensure generating a conservative envelope of the model.

\item The movement is stopped when $\mathbf{p}$ reaches $\mathcal{P}$. 
\end{itemize}
%
The conservative hull $\mathcal{C}$ is computed with the convex hull of the given model and all points of these trajectories (see Fig.\ref{fig:CompHull} for an example). To avoid potential robustness issues caused by numerical errors, we slightly enlarge $\mathcal{C}$ to compute a tetrahedral mesh $\mathcal{T}^c$. In order to ensure the compatibility of curved layers, all vertices and elements in $\mathcal{T}^m$ must be included in $\mathcal{T}^c$ (i.e., $\mathcal{T}^m \subset \mathcal{T}^c$ -- see Fig.\ref{fig:sys_pipeline}(a)). The computational domain for support is defined as $\mathcal{T}^s = \mathcal{T}^c - \mathcal{T}^m$, where the interfaces between $\mathcal{T}^s$ and $\mathcal{T}^m$ are compatible. 


%

The support layers can be generated in $\mathcal{T}^s$ by the extrapolation of $G(\mathbf{x})$ as $\Tilde{G}(\mathbf{x})=G(\mathbf{x})$ ($\forall \mathbf{x}\in \partial\mathcal{T}^m$). When using the same set of iso-values to extract the curved layers in $\mathcal{T}^m$, the compatible curved layers can be generated in $\mathcal{T}^s$ -- denoted by $\{ \mathcal{L}^s_i\}_{i=1,2,\cdots,n}$. As $\Tilde{G}(\mathbf{x})$ and $G(\mathbf{x})$ are compatible at the interface between $\mathcal{T}^m$ and $\mathcal{T}^s$, the curved layers $\{ \mathcal{L}^s_i\}$ for the support are compatible with the curved layers $\{ \mathcal{L}^m_i\}$ for the input solid.

\subsection{Overview of support generation}
The basic idea for our support generation algorithm is to form a tree-like skeleton from branches to the trunk, where the branches are used to support the overhang regions with the trunk standing on the building platform (see Fig.\ref{fig:sys_pipeline}(c) for an example). Leaf nodes of the tree are generated from the vertices on overhang faces. The nodes are progressively `projected' from the current layer to the next layer while gradually being aggregated together. After constructing this tree-like skeleton (Fig.~\ref{fig:sys_pipeline}(c)), an implicit solid is generated by the convolution surface \cite{Jin_CG02} (Fig.~\ref{fig:sys_pipeline}(d)), which is used to trim the $\{ \mathcal{L}^s_i\}$ to generate the final slimmed support for  printing curved layers (Fig.~\ref{fig:sys_pipeline}(e)).  
%
%
The specific steps include:
\begin{itemize}
  \item Generating tree-like skeleton for support (Sec.~\ref{secAlgorithm}-A);
  \item Constructing an implicit solid from the tree-like skeleton (Sec.~\ref{secAlgorithm}-B);
  \item Extracting curved layers for the support (Sec.~\ref{secAlgorithm}-C).
\end{itemize}
Details are given in the following section.

\section{Algorithms for support tree generation}\label{secAlgorithm}

\subsection{Generation of tree-like support skeleton}

\begin{wrapfigure}[10]{r}{0.3\linewidth}
\begin{center}
\hspace{-20pt}
\includegraphics[width=1.0\linewidth]{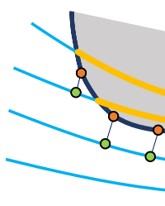}
\put(-60,60){\small \color{black}$\{\mathbf{q}^s_l\}$}
\put(-60,40){\small \color{black}$\{\mathbf{q}^s_x\}$}
\put(-70,6){\small \color{black}$\{\mathcal{L}^s_i\}$}
\put(-23,64){\small \color{black}$\{\mathcal{L}^m_i\}$}
\end{center}
\end{wrapfigure}

All of the vertices of overhang faces on the surface of model mesh $\mathcal{T}^m$ are copied as leaf nodes $\{\mathbf{q}^s_l\}$ of the support tree. A set of rays are built from $\{\mathbf{q}^s_l\}$ along the inverse direction of their corresponding $\mathbf{d}_p$ to intersect with the next curved layer below it and the intersection point is defined as $\{\mathbf{q}^s_x\}$, their branch count is initially defined as one. It should be emphasized that the overhang nodes are transferred from the model surface onto the curved layers in this step. After that, the tree tracing and merging are iteratively finished at the remaining layers below. 

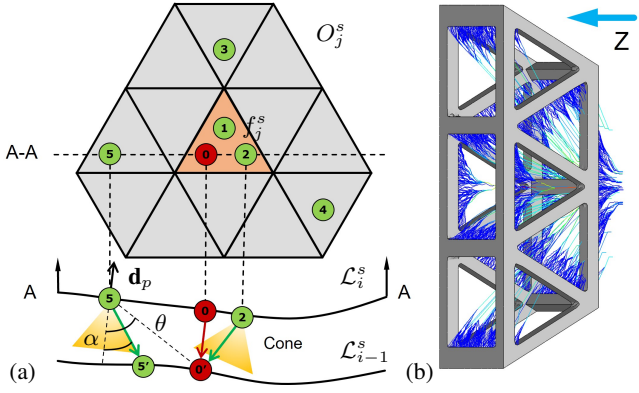
\begin{figure}[t]
\centering
\includegraphics[width=\linewidth]{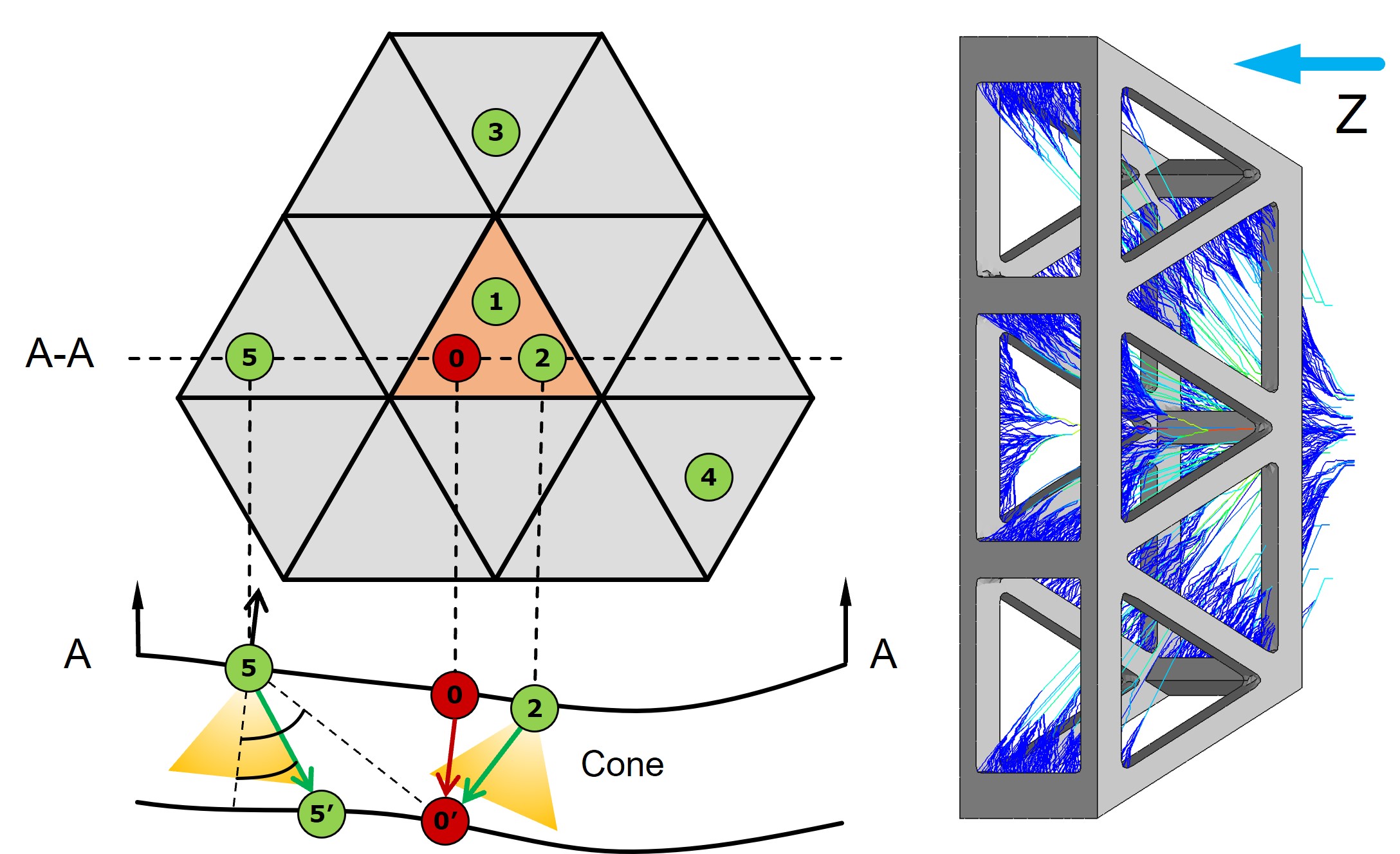}
\put(-240,2){\small \color{black}(a)}
\put(-90,2){\small \color{black}(b)}
\put(-115,38){\small \color{black}$\mathcal{L}^s_{i}$}
\put(-115,10){\small \color{black}$\mathcal{L}^s_{i-1}$}
\put(-152,95){\small \color{black}${f}^s_j$}
\put(-124,130){\small \color{black}${O}^s_j$}
\put(-185,20){\small \color{black}$\theta$}
\put(-212,15){\small \color{black}$\alpha$}
\put(-195,38){\small \color{black}$\textbf{d}_p$}
\caption{(a) Illustration of tree branch tracing and merging. “A-A” means the cross-section of $\mathcal{L}^s_{i}$, the red node is the host node and green nodes are the following nodes, while nodes with the symbol prime are the intersection nodes of tree branches on the next layer $\mathcal{L}^s_{i-1}$. Note that all the operation is conducted on the curved layers. (b) The tree skeleton for the supporting structure of the Bridge model.
}
\label{fig:branch_tracing}
\vspace{-10pt}
\end{figure}

To explain the tree skeleton generation clearly, a planar illustration is shown in Fig.~\ref{fig:branch_tracing}. There are three intersection points 
on the face ${f}^s_j$ of the layer $\mathcal{L}^s_i$. The $\textit{n}$-ring neighbour of ${f}^s_j$ is defined as ${O}^s_j$ (where the value of $\textit{n}$ will be defined below) 
and there are other three intersection nodes on the ${O}^s_j$. Three operations are conducted in each iteration:

\begin{enumerate}
    \item The intersection point that has the greatest amount of tree branches in the ${f}^s_j$ is defined as host node $\mathbf{q}^s_h$ on the tree (refer to the red node `0' in Fig.~\ref{fig:branch_tracing}). Then it is projected directly along the inverse growing direction which is decided by the face normal of ${f}^s_j$ and the intersection point is called $\mathbf{q'}^s_h$ on the tree (node `$0'$').
    \item The remaining intersection points in ${f}^s_j$ and ${O}^s_j$ are treated as follower nodes $\mathbf{q}^s_f$, and the branches growing from them are rotated towards $\mathbf{q'}^s_h$ with target angle $\theta$. In Fig.~\ref{fig:branch_tracing}, the following nodes marked with “2” and “5” shoot green arrows toward the below layer $\mathcal{L}^s_{i-1}$.
    \begin{itemize}
        \item Case $\alpha \le \theta$: The arrow rotates with the maximum allowed angle $\alpha$ and intersects with $\mathcal{L}^s_{i-1}$ at node “$5'$”;
        \item Case $\alpha > \theta$: The arrow starting from intersection node “2” can directly point to “$0'$” in the layer $\mathcal{L}^s_{i-1}$ and the end point of arrows shot both from host node “0” and following node “2” are merged into “$0'$” and the branch amount of node “2” is absorbed by “$0'$”.
    \end{itemize}
    \item Each edge whose start node is on the layer $\mathcal{L}^s_i$ and the end node is on the layer $\mathcal{L}^s_{i-1}$ is constructed and  added into a tree graph. 
\end{enumerate}
 
Note that the support structure itself should be support-free according to the predefined curved layers obtained from $\Tilde{G}(\mathbf{x})$;  
in other words, it should support itself before supporting the model. This requirement is translated as the rotation angles of branches
should be less than $\alpha$,  where $\alpha$ is the self-supporting angle. Hence, rotation arrows should not be out of the orange cone, as shown in Fig.~\ref{fig:branch_tracing}. The size of $\textit{n}$-ring neighbor means how many rings of neighbor faces are collected for the host node and the default number of rings is 3 in our system. The iteration of tracing is stopped when the nodes reach the platform or model layers.

\subsection{Implicit solid construction based on tree-skeleton}

\begin{figure}[t]
\centering
\includegraphics[width=\linewidth]{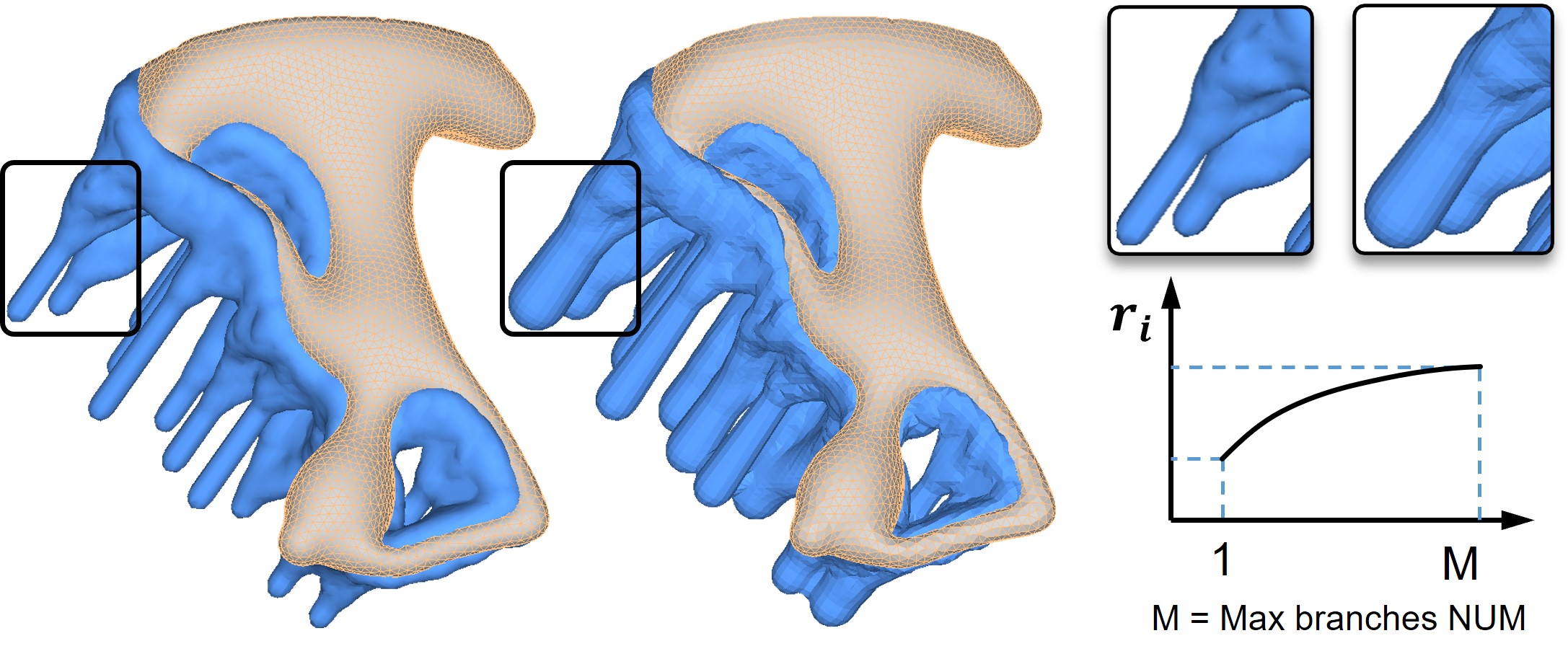}
\put(-240,6){\small \color{black}(a)}
\put(-160,6){\small \color{black}(b)}
\vspace{-5pt}
\caption{Comparison of the implicit surface of support structure building with (a) fixed branch radius; (b) dynamic branch radius. The latter one is becoming gradually thicker from the top of the tree to the major trunk.}
\label{fig:implicit_surface}
\end{figure}

\begin{figure*}[t]
\centering 
\includegraphics[width=\linewidth]{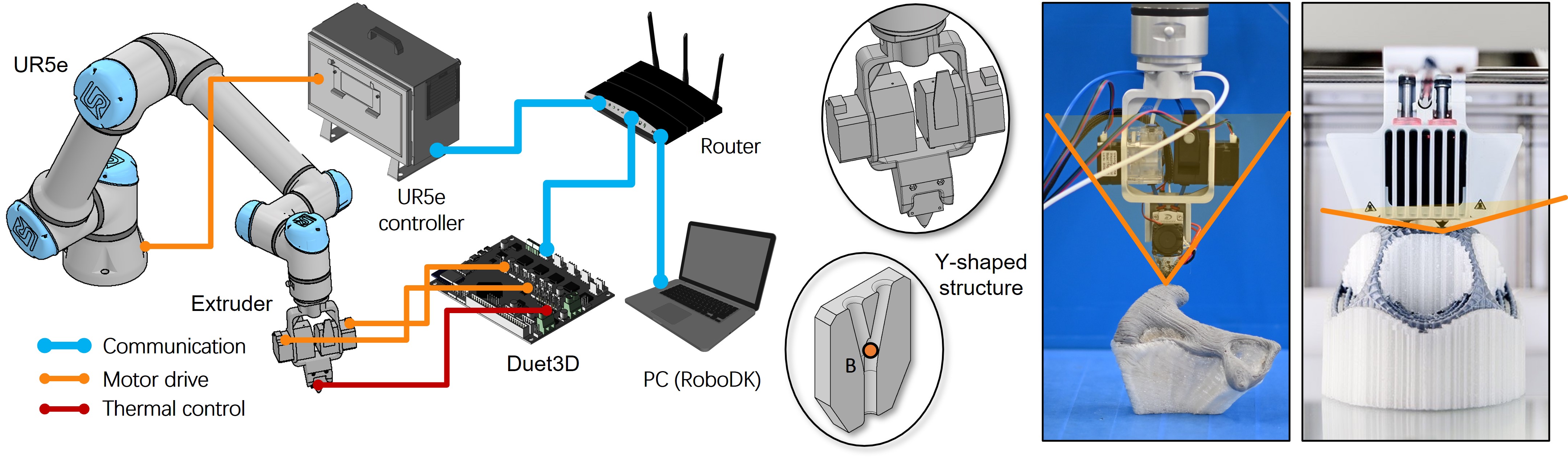}
\put(-505,10){\small \color{black}(a)}
\put(-165,10){\small \color{black}(b)}
\put(-80,10){\small \color{black}(c)}
\vspace{-10pt}
\caption{(a) The hardware of the 3D printing system for FDM-based dual-material fabrication. A 2-in-1  extruder is installed on the end-effector of the robot arm, different filaments can be controlled to pass through the Y-shape structure alternatively. A router is employed as the core of communication between the controller of UR5e, Duet3D, and the laptop PC. (b) and (c) illustrate the convex cone (orange) formed by the extruder of our setup and Ultimaker, it is easy to find that ours has a sharper envelope and can avoid local collision more easily during 3D printing.}
\label{figSysHardware}
\vspace{-5pt}
\end{figure*}

The implicit solid is built from the tree skeleton and used to extract slimmed layers of support from $\{\mathcal{L}^s_i\}$. As the extraction operation will be 
conducted between layer mesh and implicit solid, it is robust and time-saving compared to the trimming operation between meshes. The tree skeleton $\Omega$ is represented as a complex $\Omega=(\mathcal{V},\mathcal{E})$ with a set of vertices and edges. Each $v_i\in\mathcal{V}$ defines the position of vertices, and each edge $e_j\in\mathcal{E}$ is represented as a pair of vertices associated with the radius of the edge’s corresponding strut as $e_j=(v_s,v_e,r_j)$. The implicit solid is defined around the tree skeleton and formulated as
\begin{equation}
\mathcal{S}(\Omega)=\{\mathbf{p} | F (\mathbf{p}) \ge 0 (\forall \mathbf{p} \in\Re^3)\},
\end{equation}
where $F(\cdot)$ is an implicit field value at query point $\mathbf{p}$, 
and it can be written in the edge form as
\begin{equation}
F(\mathbf{p})=-C+ \sum_{e_j \in \Omega} r_j \int_{\mathbf{x} \in e_j } f(\mathbf{p}-\mathbf{x}) d e_j,
\label{implicit solid equation}
\end{equation}
where $C$ is a constant iso-value defined according to the radii defined on the skeleton edges and the quartic polynomial kernel function is adopted. 
The implicit function value at point $\mathbf{p}$ contributed by $e_j$ can be computed as
\begin{equation}
\begin{aligned}
    F_{e_j} (\mathbf{p}) &= r_j \int_{s_1}^{s_2} (1-\frac{\| \mathbf{p}-\mathbf{x(s)} \|^2}{R^2})ds\\&= \frac{r_j}{15R^4} (3l^4 s^5-15al^2 s^4+20a^2 s^3 )\|_{s_1}^{s_2},
\end{aligned}
\end{equation}
where $R$ is the support size and the definition of other terms can be found at \cite{Liu_21JCISE}. 
When the value of $r_j$ is decreased, the implicit solid boundary moves toward $e_j$ and thus branch radius is reduced. In our algorithm, the value of each $r_j$ is decided by the branch count which is recorded during the tree skeleton building. Inspired by the trees in the natural environment, the sum of the cross-section area of all branches should be equal to the area of the trunk with the radius $r$. Therefore, we have
\begin{equation}
    \pi r_j^2 = \sum^m_{k = 1} \pi r_k^2,
\end{equation}
where the $m$ is the number of branches. This calculation method of branch radius can obtain a support tree with thick trunks and slim branches as Fig.\ref{fig:implicit_surface}(b) shows. However, the truck part of  Fig.\ref{fig:implicit_surface}(a) is too weak to support the model itself. 

\subsection{Extracting curved layers of support}
\begin{figure}[t]
\centering
\includegraphics[width=\linewidth]{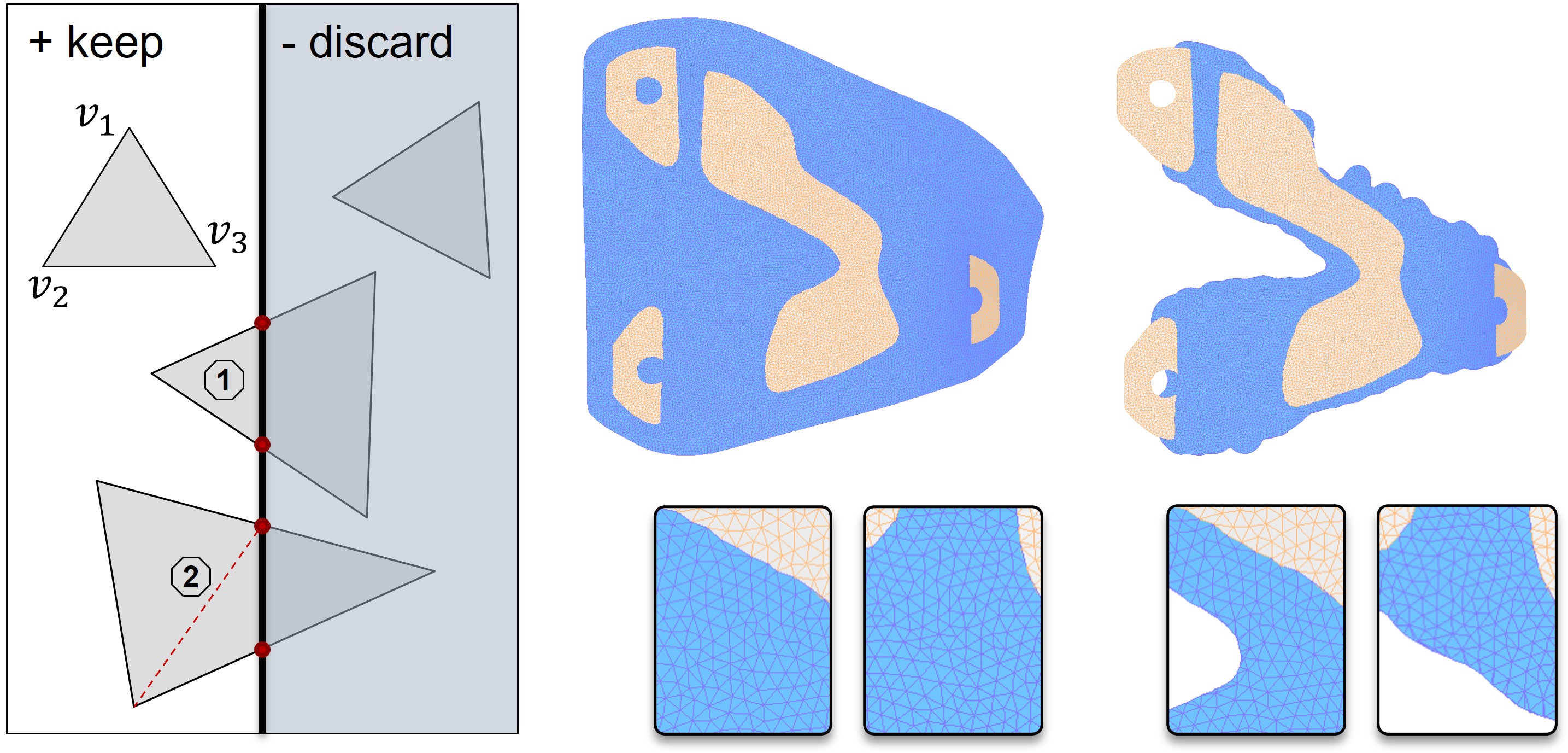}
\put(-243,5){\small \color{black}(a)}
\put(-156,5){\small \color{black}(b)}
\put(-75,5){\small \color{black}(c)}
\caption{Illustration of slimmed tree-like support generation. (a) Cases for conducting trimming operation; (b) The compatible layers (support layer and model layer) before trimming; (c) The rest support layer after extraction and the original model layer. 
}
\label{fig:tree_support_extraction}
\vspace{-15pt}
\end{figure}

In this section, the extraction operation between support layers $\{\mathcal{L}^s_i\}$ (as a set of polygons extracted as iso-surface of $\Tilde{G}(\mathbf{x})$) and implicit solid $\mathcal{S}(\Omega)$ is described to obtain the final slimmed support layers $\{\mathcal{L'}^s_i\}$. 
In Fig.~\ref{fig:tree_support_extraction}, $v_i (i = 1,2,3)$ represent vertices of the face $f_j$ on a support layer and their implicit function values ${F(v_i)}$ are computed. The number of vertices with ${F(v_i)}>0$ is defined as $N$, and so there are four cases:
\begin{itemize}
    \item $N = 0$: If implicit field values of all vertices are less than or equal to zero, then the facet will be discarded from slimmed support layers.
    \item $N = 1$: $\mathcal{S}(\Omega)$ will cut the facet and keep the small triangle formed by two intersection vertices and the original vertex (see the marker 1 in Fig.~\ref{fig:tree_support_extraction}(a)).
    \item $N = 2$: $\mathcal{S}(\Omega)$ will pass through the facet and the quadrilateral part will be collected. Furthermore, the quadrilateral is split into two triangles (refer to marker 2 in Fig.~\ref{fig:tree_support_extraction}(a)).
    \item $N = 3$: If implicit field values of three vertices on $f_j$ are all larger than zero, then $f_j$ is kept.
\end{itemize}

After extracting the trimmed triangles, the final mesh surface for a curved layer of support is obtained. Contour parallel toolpaths~\cite{Fang_SIG20} are generated on the curved layer to complete the robot-assisted 3D printing.

\section{Implementation Details and Results}\label{secResult}

\subsection{Hardware implementation}

The core of our systems is  mainly composed of a UR5e robotic arm equipped with a 2-in-1 extrusion system as shown in Fig.~\ref{figSysHardware}. The extrusion system includes a Y-shaped structure that can automatically switch between two different materials. In our implementation, \textit{polylactic acid} (PLA) and \textit{polyvinyl alcohol} (PVA) are used to print the model and the supporting structure respectively. The currently used filament is retracted to the bifurcation point $B$ of the Y-shape mechanism, and the other filament enters the nozzle through point B when switching the material. 
A custom made cone-shape heater block is used to decrease the collision possibility when printing curved layers
(as shown in Fig.~\ref{figSysHardware}(b) and (c)).

Compared to the configuration of \cite{Wu_ICRA17} already discussed in the related work, our extruder system is directly installed on the end effector of the UR5e, so we can fully utilize the precision of the robot and the calibration method of \textit{tool center point} (TCP) provided by the UR robot. This setup makes the calibration process simpler and can achieve high precision for the end effector more easily.
The control part of our system is based on the software  
RoboDK~\cite{RoboDK_web}. 
A Duet3D control board~\cite{Duet3D_web} is used to control the temperature of the nozzle, drive the extrusion motor for feeding materials, and control the fans for cooling. The controller 
of UR5e, the Duet3D board and the laptop PC running RoboDK are linked with each other by a router. The UR5e control box dispatches the motion commands and the extrusion commands to the robot and the extruder simultaneously.


\subsection{Computational results and physical experiments}

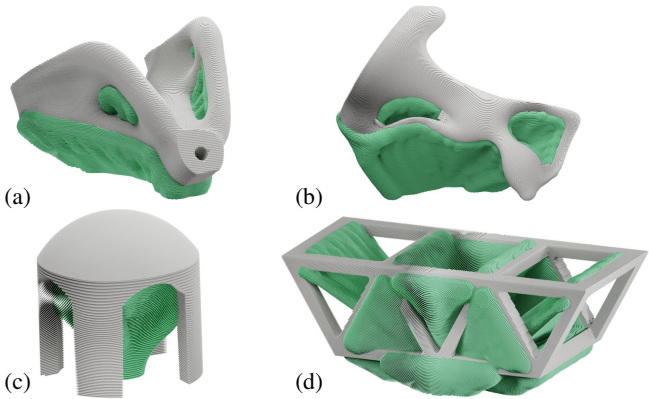
\begin{figure}[t]
\includegraphics[width=\linewidth]{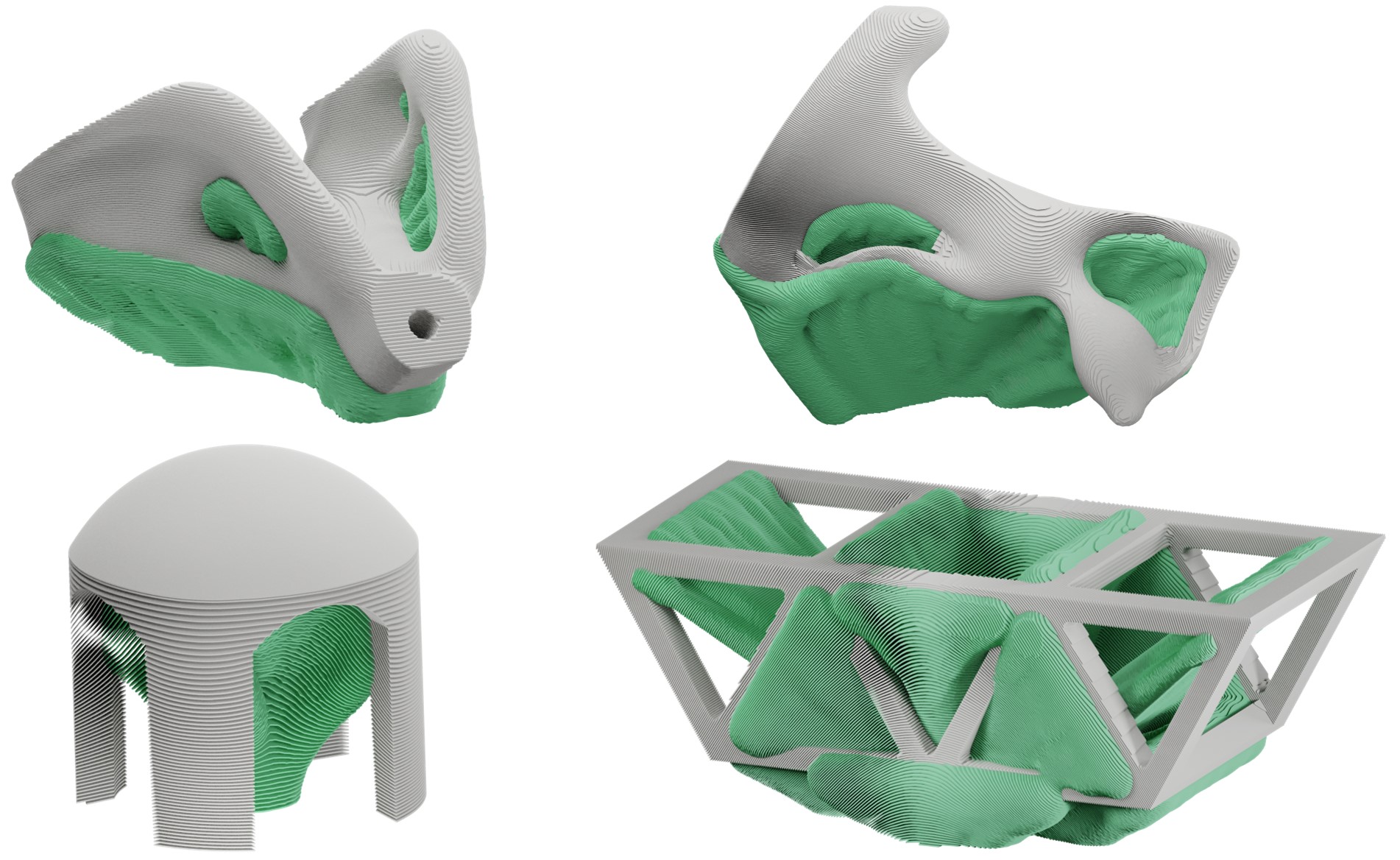}
\put(-246,75){\small \color{black}(a)}
\put(-136,75){\small \color{black}(b)}
\put(-246,5){\small \color{black}(c)}
\put(-136,5){\small \color{black}(d)}
\vspace{-5pt}
\caption{Slicing results of different examples that are supported by proposed method (a) Topo-Opt, (b) Yoga, (c)  Dome, and (d) Bridge.}
\label{fig:slicing_results}
\end{figure}

The algorithm presented above is implemented in a C++ program to generate a slimmed support structure for 3D printing with curved layers. The numerical solver Eigen \cite{eigenweb} is employed to solve linear systems. 

We have tested the slimmed support generation method on a variety of models.  The first example of our tests, shown in Fig.~\ref{fig:sys_pipeline}, is a Dome model. Specifically, the middle part of the Dome model is well-supported and the support structure is slimmed into a tree-like shape. Fig.~\ref{fig:branch_tracing}(b) displays the tree-like skeleton of the Bridge model and Fig.~\ref{fig:implicit_surface} illustrates the implicit solid generated from the tree-like skeleton of a Yoga model. The slicing results of the four models are shown in Fig.~\ref{fig:slicing_results} where the grey part denotes the layers for the main model and the green part is for the curved layers of support. 



\begin{figure}[t]
\includegraphics[width=\linewidth]{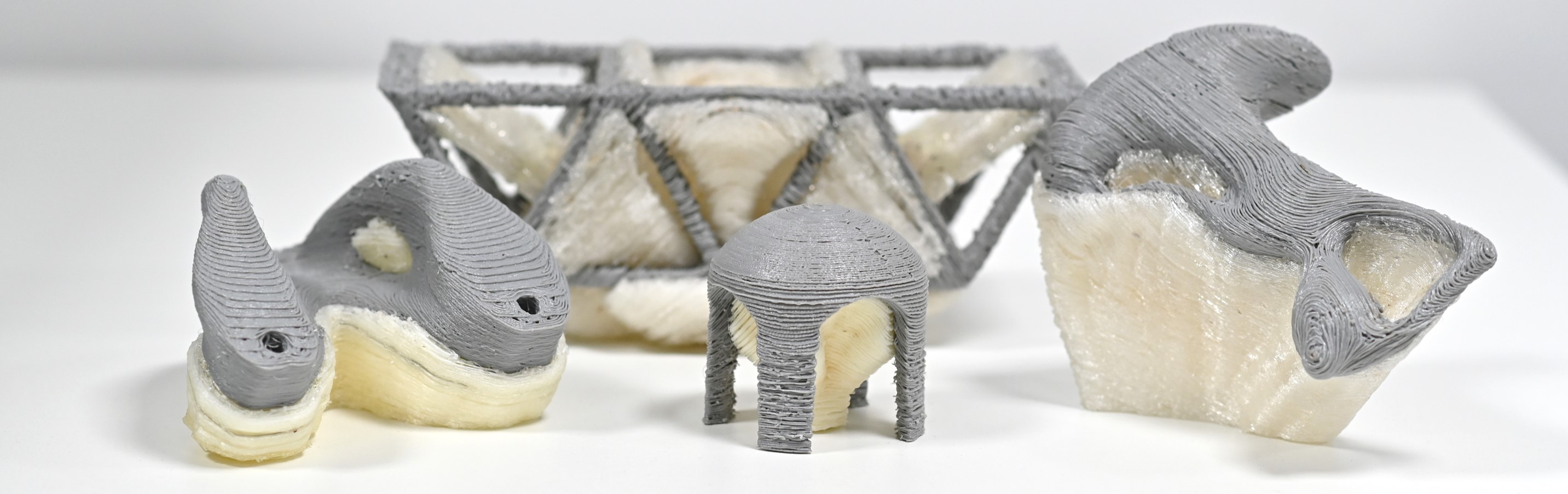}
\caption{Example models that have been fabricated by our robotic system -- (left) Topo-Opt, (middle) Dome, and (right) Yoga with supporting structures.}
\label{fig:experiments}
\vspace{-10pt}
\end{figure}

The printing volumes compared between our method and~\cite{Fang_SIG20} are shown in Table~\ref{tab:comparison_support_volume}. Parameters such as the number of layers, nozzle diameter (0.8mm), toolpath pattern, etc. are the same as each other in the tests. The comparison of final printing results is also given in Fig.~\ref{figTeaser}(c) and (d). The reduction in the volumes of the supports ranges from 32.5\% to 59.7\%.
More experimental tests have been conducted to verify the performance of our support generation method (see Fig.~\ref{fig:experiments}). The statistics of computation and fabrication are given in Table~\ref{tab:compute_and_physical_statistic} and the computational time is much less than printing experiments. Finally, the process of physical printing can be found in the supplementary video of this paper.

\begin{table}[t]
\caption{Comparison of support volume}\vspace{-5pt}
\centering\label{tab:comparison_support_volume}
\footnotesize
\begin{tabular}{r||c|c|c }
\hline
Sup. Vol. ($mm^3$) &  Previous Method~\cite{Fang_SIG20}      &  Current & Reduction \\
\hline \hline
Topo-Opt   &  $36,065.1$ &  $24,340.7$  &  $\downarrow 32.5\%$\\
Yoga       &  $31,339.6$ &  $17,675.5$  &  $\downarrow 43.6\%$\\
Dome       &  $9,953.2$  &  $7,434.9$   &  $\downarrow 25.3\%$\\
Bridge     &  $211,477.4$&  $87,340.2$  &  $\downarrow 59.7\%$\\
\hline
\end{tabular}
\end{table}

\begin{table}[t]
\caption{Statistics of compute and physical fabrication.}\vspace{-5pt}
\centering\label{tab:compute_and_physical_statistic}
\footnotesize
\begin{tabular}{r||c|c|c|c|r }
\hline
&\multicolumn{3}{c|}{Compute (sec.)} & \multicolumn{2}{c}{Fabrication}\\
\cline{2-6}
Model &  Tree Gen. & Trimming & Toolpath & Layer \# $^\dag$  & Time   \\
\hline \hline
Topo-Opt   &  $63.5$ &  $533.1$  &  $63.8$  & $100/137$  &  $13.6$h   \\
Yoga       &  $40.1$ &  $296.5$  &  $10.4$  & $160/177$  &  $10.9$h   \\
Dome       &  $12.5$ &  $79.3 $  &  $4.8$   & $80/92$    &  $3.5 $h   \\
Bridge     &  $52.1$ &  $668.8$  &  $15.4$  & $150/166$  &  $32.8$h   \\
\hline
\end{tabular}
\begin{flushleft}
$^\dag$~The model layer and support layer number of each model.
\end{flushleft}
\end{table}

\section{Conclusion and Discussion}\label{secConclusion}
In this paper, a skeleton-based support generation method is presented for robot-assisted 3D printing with curved layers. 
Since the solids of support are represented as implicit surfaces defined by the skeletons, the problems of numerical robustness can be effectively avoided when extracting the curved layers for supporting structures. The curved layers of support are also compatible with the curved layers of the primary model. We have verified the effectiveness of our algorithm on a dual-material printing platform that uses a robotic arm and a newly designed dual-material extruder. Our experimental tests give very encouraging results, and the models with large overhangs can be well fabricated with supporting structures generated by our method. 

There are still limitations in our current implementation of the support generation method. The discontinuity between separated regions on the curved layers of supporting structures will lead to repeated retraction operations that have an influence on the quality of 3D printing. We plan to further optimize the skeleton of support to improve this in future work. Moreover, the stability of the partially printed model needs to be considered when being applied to hardware with table-tiling configuration \cite{Fang_SIG20}. Some interesting issues also need to be explored, such as the influence of the initial selection of host nodes, the size of neighbor rings, and the distribution of branches to facilitate the removal of supports.

\bibliographystyle{IEEEtran}
\bibliography{ICRA23}

\end{document}